# Title

Agent-Based Uncertainty Awareness Improves Automated Radiology Report Labeling with an Open-Source Large Language Model

## Authors

1. Hadas Ben-Atya (MS), Faculty of Biomedical Engineering, Technion - Israel Institute of Technology, Haifa, Israel.

2. Naama Gavrielov, Faculty of Biomedical Engineering, Technion - Israel Institute of Technology, Haifa, Israel.

3. Zvi Badash, Faculty of Data and Decision Sciences, Technion - Israel Institute of Technology, Haifa, Israel.

4. Gili Focht (MSc), Juliet Keidan Institute of Pediatric Gastroenterology Hepatology and Nutrition, Shaare Zedek Medical Center, The Hebrew University School of Medicine, Jerusalem, Israel.

5. Ruth Cytter-Kuint (MD), Pediatric Radiology Unit, Radiology Department, The Eisenberg R&D Authority, Shaare Zedek Medical Center, The Hebrew University of Jerusalem, Jerusalem, Israel.

6. Talar Hagopian (MD), Pediatric Radiology Unit, Radiology Department, The Eisenberg R&D Authority, Shaare Zedek Medical Center, The Hebrew University of Jerusalem, Jerusalem, Israel.

7. Dan Turner (MD PhD), Juliet Keidan Institute of Pediatric Gastroenterology Hepatology and Nutrition, Shaare Zedek Medical Center, The Hebrew University School of Medicine, Jerusalem, Israel.

8. Moti Freiman (PhD), Faculty of Biomedical Engineering, Technion - Israel Institute of Technology, Haifa, Israel.

## Corresponding author contact information

E-mail address: moti.freiman@technion.ac.il

Full address:  Technion, Faculty of Biomedical Engineering, Silver Building, 3200001

**Funding information**

The study was sponsored by the Leona M. and Harry B. Helmsley Charitable Trust. The funders had no role in the study design, data collection and analysis, decision to publish, or preparation of the manuscript.

**Abstract:**

**Purpose:** To improve the reliability and performance of Large Language Models (LLMs) in extracting structured data from radiology reports, particularly in domains with complex and non-English texts (e.g., Hebrew), by incorporating agent-based uncertainty-awareness method to achieve trustworthy predictions in medical applications.

**Materials and Methods:**

This retrospective study analyzed 9,683 Hebrew radiology reports from Crohn's disease patients (2010–2023) across three medical centers. A subset of 512 reports was manually annotated for six gastrointestinal organs and 15 pathological findings, while the remaining data was automatically annotated using the HSMP-BERT model.

Structured data extraction used Llama 3.1 (Llama 3-8b-instruct), an open-source LLM, with Bayesian Prompt Ensembles (BayesPE) leveraging six semantically equivalent prompts for uncertainty estimation. An Agent-Based Decision Model synthesized multiple prompt outputs into five confidence levels for calibrated uncertainty, compared against three entropy-based models. Performance was assessed via accuracy, F1 score, precision, recall, and Cohen's Kappa, both before and after filtering high-uncertainty cases.

**Results:** The agent-based model outperformed the baseline across all metrics, achieving an F1 score of 0.3967, recall of 0.6437, and Cohen's Kappa of 0.3006. After filtering high uncertainty cases ($\geq 0.5$), the F1 score increased to 0.4787 and Kappa to 0.4258. Uncertainty histograms showed clear separation between correct and incorrect predictions, with the agent-based model exhibiting the best-calibrated uncertainty predictions.

**Conclusion:** Incorporating uncertainty-aware prompt ensembles and an agent-based decision model significantly enhances the performance and reliability of LLMs in structured data extraction from radiology reports, offering a calibrated and interpretable approach for high-stakes medical applications.

## Summary Statement:

An uncertainty-aware approach using prompt ensembles and an agent-based decision model improved the reliability, interpretability, and performance of LLMs in extracting structured data from radiology reports, with calibrated predictions and enhanced metrics.

## Key Points:

- Bayesian prompt ensembles improved accuracy and reliability in extracting structured data from Hebrew radiology reports, with the agent-based decision model achieving top performance (F1: 0.4787, Kappa: 0.4258).
- Uncertainty-aware filtering enhanced metrics and clearly distinguished correct from incorrect predictions, as shown by calibrated uncertainty histograms.
- The agent-based decision model delivered robust, interpretable results by synthesizing multiple prompts and categorizing decisions by confidence levels.

## Introduction:

Radiology reports contain critical clinical information, essential for healthcare decision-making, retrospective studies, and radiologic image annotation. Despite ongoing efforts to implement structured reporting [1], unstructured free-text reports remain dominant, posing challenges for large-scale data analysis due to their variability and lack of standardization. Automating the extraction of structured data addresses these issues by streamlining workflows, reducing manual effort, and improving data consistency and accessibility. This approach not only enhances patient care but also enables large-scale research through comprehensive meta-analyses, accelerated scientific discoveries, and stronger evidence-based practice [2-4].

Large language models (LLMs) can automate structured data extraction from radiology reports by processing unstructured text to efficiently yield clinically relevant insights [5-7]. However, privacy concerns around sensitive medical data have prompted interest in open-source LLMs, which offer greater transparency, adaptability, and control. Models like Llama 3.1 can extract meaningful insights from medical texts while mitigating risks associated with proprietary systems [8].

Despite their potential, applying LLMs to radiology remains challenging. Overconfidence in predictions undermines reliability in high-stakes medical contexts [9], highlighting the need for robust methods to assess and manage uncertainty. Additional obstacles arise in underrepresented languages like Hebrew, where limited high-quality training data hampers performance. Innovative strategies to enhance LLM adaptability and reliability are essential for effective deployment across diverse linguistic and clinical settings.

Recent progress in uncertainty quantification enhances LLM reliability and interpretability in high-stakes domains like healthcare and radiology. Techniques such as Confidence Elicitation (CE) [10-12], Token-Level Probabilities (TLP), and Sample Consistency (SC) address overconfidence by providing uncertainty measures. However, these approaches treat all responses with equal weight may overlook variations in quality, relevance, or coherence. Refining these methods is crucial for meeting the demands of increasingly complex applications.

Bayesian Prompt Ensembles (BayesPE) [13] enhance uncertainty estimation in LLMs by combining semantically equivalent prompts and optimizing their weights via variational inference, improving prediction reliability and calibration. However, its reliance on a parametric model for prompt weights may limit flexibility in complex or dynamic settings.

Agent-based approaches [14] offer adaptive decision-making for LLM outputs, potentially addressing BayesPE's limitations by incorporating more sophisticated uncertainty estimation processes. This study aims, therefore, to introduce and evaluate an agent-based method for uncertainty-aware, LLM-based structured data extraction from real-world Hebrew radiology reports of Crohn's disease patients, comparing its performance and robustness to the probabilistic aggregation methods of BayesPE.

## Materials and Methods:

### Data Collection

The retrospective multicenter study was approved by the institutional review board. Informed consent was waived due to the retrospective nature of the study.

Free-text Hebrew radiology reports of Crohn's disease patients were sourced from the EPI-IIRN study [15]. The dataset comprises 9,683 reports, each corresponding to a patient visit, collected from 8,093 unique patients across three medical institutions. These reports document Magnetic Resonance Enterography or CT Enterography exams conducted between 2010 and 2023.

### Manual Data Annotation

We randomly selected 512 reports from the dataset, each manually annotated by a trained radiologist (T.H.) for six gastrointestinal organs (Jejunum, Ileum, Cecum, Colon, Sigmoid, and Rectum). Each report included up to 15 distinct pathological findings (e.g., Inflammation, Wall thickening, Ulceration) [16], resulting in 90 organ–finding combinations. Labels were assigned as 0 (negative), 1 (positive), 2 (organ absent due to surgery), or 9 (organ not visible on MRI). In the main experiment, we treated labels 2 and 9 as negative, leaving two possible outcomes per organ–finding combination. We excluded combinations with ≤15 positive cases, yielding 23 final organ–finding combinations. Fig. 1 shows the distribution of all organ-specific findings and the subset with at least 15 positive samples. Detailed information about the prevalence of the different labels in the test set is provided in Table S2 in the supplementary materials.

**NLP Based Annotations**

We used HSMP-BERT, an in-house BERT-based NLP model for structured data extraction from Hebrew radiology reports of Crohn's disease patients [17,18], to annotate the entire dataset of 9,683 reports. The model was trained on the manually annotated subset, focusing on 23 prevalent organ–finding combinations. Performance metrics for these labels appear in Appendix A. To ensure high-quality data for the BayesPE-based methods, we retained only labels with a Cohen's Kappa score above 0.7, resulting in eight final labels: Ileum comb sign, Ileum inflammation, Ileum pre-stenotic dilation, Ileum stenosis, Ileum wall enhancement, Ileum wall thickness, Rectum wall thickness, and Sigmoid comb sign.

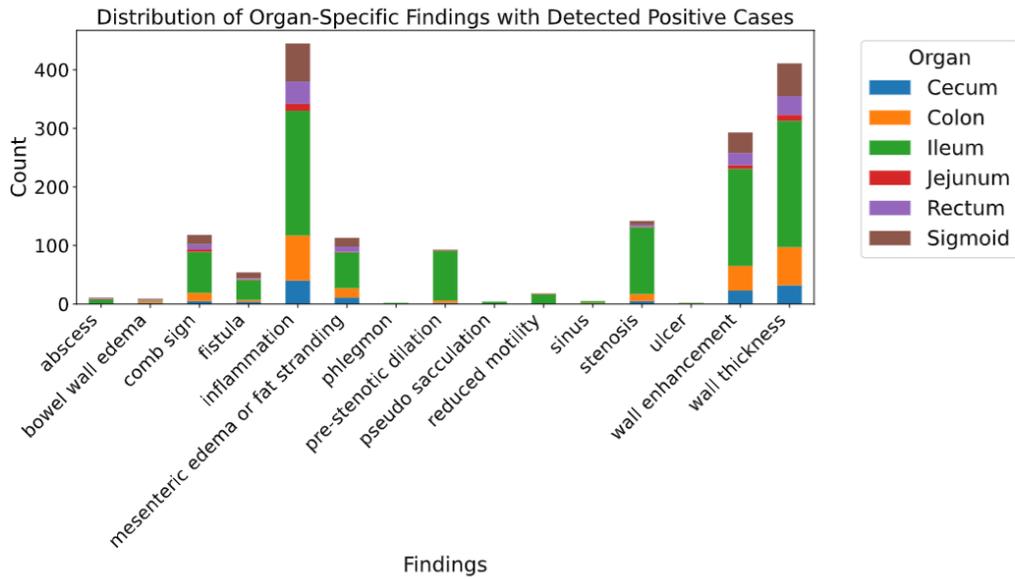

(a) Distribution of organ-specific findings with detected positive cases.

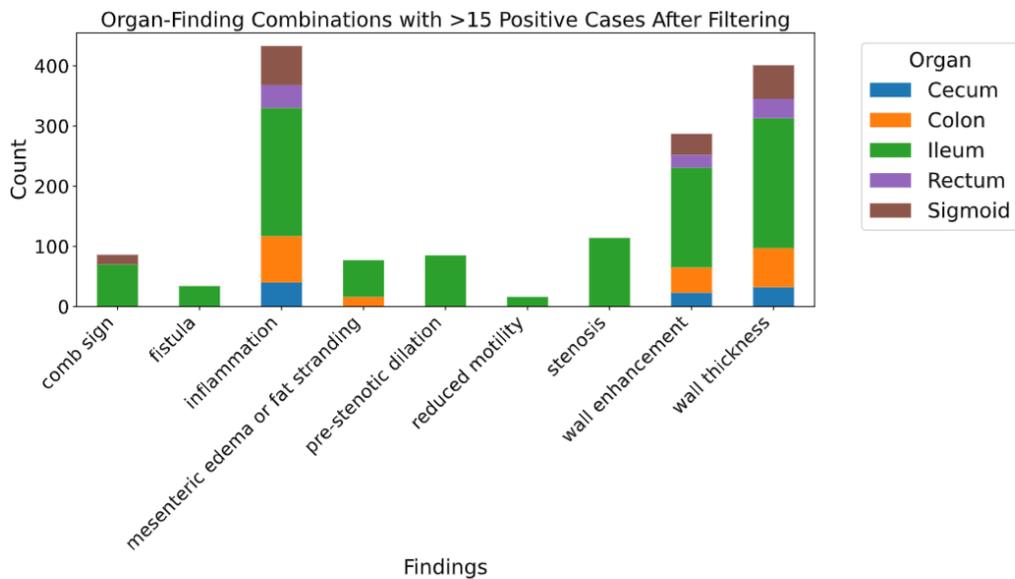

(b) Organ-finding combinations with 15+ positive cases after filtering.

**Fig. 1:** (a) Distribution of organ-specific findings and (b) filtered organ-finding combinations with more than 15 positive cases.

**Large Language Model (LLM) utilization for structured data extraction**

We used the Llama 3.1 model (Llama 3-8b-instruct) [19], part of Meta's multilingual LLM collection, to extract structured data from Hebrew radiology reports. Although it supports several languages (English, German, French, Italian, Portuguese, Hindi, Spanish, Thai), Hebrew is not among them. Llama 3.1 utilizes an auto-regressive transformer architecture [20], further refined via supervised fine-tuning (SFT) [21] and reinforcement learning from human feedback (RLHF) [22-24] to better align with human preferences. Given its limited Hebrew support, we carefully adapted the model for this context.

We prompted the model with various organ–finding combinations to extract structured data from Hebrew radiology reports, asking whether a specific finding was indicated. We also included a self-explanation step for each answer, aiding model validation and improving accuracy, as previously demonstrated. Specifically, we used the following prompt:

> **LLM Prompt example:**
>
> "Does the following radiology report indicate that the patient has {label}? Here is the report: {{}}. Please provide a concise response in JSON format, adhering to the schema: {{'Answer': 'Yes/No', 'Explanation': 'str'}}. Important: Only return a single valid JSON response. For instance: {{'Answer': 'Yes', 'Explanation': 'The report states that the patient has a severe case of the disease.'}}"

**Uncertainty Estimation with Bayesian Prompt Ensembles**

We employed the Bayesian Prompt Ensembles (BayesPE) method [13] to estimate uncertainty in LLM-generated predictions by leveraging an ensemble of semantically equivalent prompts, without modifying the LLM's architecture or requiring retraining. In alignment with their findings, we selected six prompts to balance the benefits of prompt diversity with computational efficiency, as adding more prompts provided only marginal gains. We used ChatGPT [25] to design these six prompts, each with identical semantic intent to determine the presence of specific findings in radiology reports. This ensures consistency in task execution despite variations in wording. The complete prompt structures appear in Appendix B.

We developed an agent-based method and three entropy-based approaches to quantify model uncertainty using outputs from multiple prompts. Fig. 2 illustrates the different approaches we utilized to assess uncertainty.

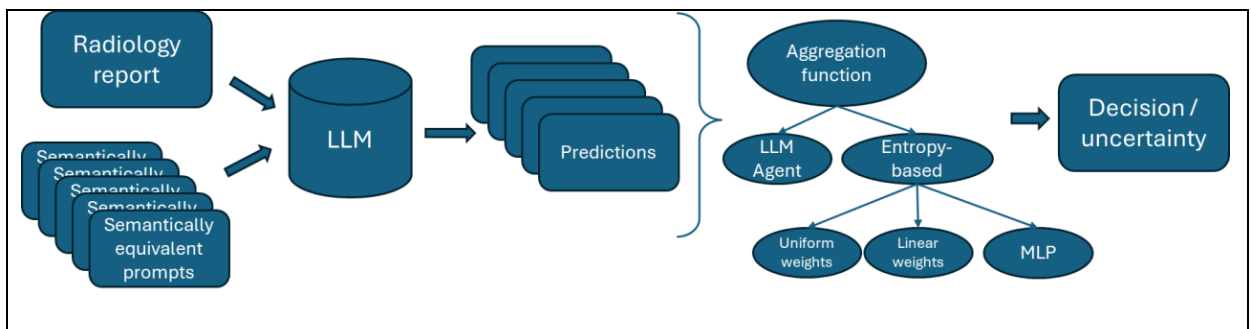

**Fig. 2:** Illustration of the proposed Bayesian Prompt Ensemble pipeline for uncertainty-aware predictions. A radiology report and multiple semantically equivalent prompts are fed into an LLM, generating one prediction per prompt. These predictions are then aggregated to yield a final decision and uncertainty estimation. The aggregation function can be either an LLM agent or an entropy-based function that applies uniform weights, learned fixed weights, or weights predicted by a pre-trained MLP model.

1. <u>Agent Decision Model</u>

    We propose an Agent Decision Model to consolidate the outputs of multiple prompts, derive a final decision, and quantify its uncertainty. This model synthesizes responses and explanations from an ensemble of prompts, producing a unified decision with a rationale for its conclusion. The agent evaluates response consistency, assesses the clarity and coherence of explanations, and identifies indications of uncertainty. Based on these factors, it categorizes the final decision into one of five distinct confidence levels:

- Definitely Yes,
- Likely Yes,
- Definitely No,
- Likely No,
- Uncertain.

To ensure interpretability and structure, the agent outputs its decision and reasoning in a predefined JSON format: {"Decision": "str","Explanation": "str"\}. This formatting requirement ensures that the output is consistent and machine-readable. The agent receives as input a set of answers ("Yes" or "No") and their corresponding explanations generated by the LLM with the different prompts. The agent evaluates these inputs based on the consistency of the answers, the clarity of their explanations, and the degree of ambiguity to produce its final decision. In our implementation, we utilized the Llama 3 - 70B model as the agent. Its outputs include:

- A final decision (e.g., "Definitely Yes", "Likely No").
- An explanation for the decision, highlighting the key factors that influenced its reasoning.

The agent's uncertainty level is derived based on the confidence category of its decision:

- Definitely Yes/No: Uncertainty = 0
- Likely Yes/No: Uncertainty = 0.5
- Uncertain: Uncertainty = 1

This semi-quantitative approach enables the seamless integration of uncertainty measures into downstream analyses, enhancing the interpretability of the decision-making process. For the "Uncertain" response, a final "Yes" or "No" decision is determined by aggregating the probabilities for each option from the previous step and selecting the option with the highest overall score.

An example of two responses by the backbone LLM that are provided as input to the agent and an example of the agent output for the input responses presented are provided below.

**Agent inputs (2 examples of the LLM output):**

- **Answer:** "No".
  **Explanation:** "The report does not mention Cecum inflammation, but rather mentions air-filled bubbles in the soft tissue of the right abdomen, likely originating from the small intestine."
- **Answer: "Yes".**
  **Explanation:** "The report mentions the presence of bubbles or air pockets in the soft tissue of the right abdomen, which may be indicative of inflammation in the cecum."

**Agent output:**

- **"Decision":** "Likely No",
  **"Explanation":** "The majority of the answers (3 out of 4) suggest that there is no inflammation in the Cecum, and the explanations provided are clear and consistent. The one answer that suggests 'yes' is not strongly supported and appears to be an over-interpretation of the report."

2. <u>Entropy-based Decision Models</u>

Alongside our agent-based decision model, we implemented three entropy-based models for comparison. These models determine the final prediction and its entropy by applying different weighting methods to the prompt outputs, as follows:

a) Uniform weights.
b) Linearly optimized weights.
c) Learnable weights using MLP.

We describe each in detail below.

**Uniform Weights:** In this approach, all prompts are assigned equal weights:

$$w_i = \frac{1}{\text{len(Prompts)}}$$

This method operates under the assumption of no prior knowledge regarding the efficacy of individual prompts, thus distributing trust uniformly across all prompts. Such an approach enhances simplicity and mitigates potential biases in weight assignment.

**Linearly Optimized Weights:** In this approach, we optimized the weights assigned to different prompts using a small validation set. We determine the weights by minimizing the following objective function:

$$L(w_{\text{raw}}, \log(P(y^*|a_i, x))) = \sum_{i=1}^{N} w_i \cdot \log(P(y^*|a_i, x)) - \sum_{i=1}^{N} w_i \log(w_i)$$

where $y^*$ denotes the correct answer (e.g., "Yes" or "No"), $a_i$ represents the $i^{th}$ prompt structure, and $x$ is the radiology report. The weights $w_i$ are computed by applying a softmax transformation to the raw weights $w_{\text{raw}}$:

$$w_i = \frac{\exp(w_{\text{raw},i})}{\sum_{j=1}^{N} \exp(w_{\text{raw},j})}$$

The first term $\sum_{i=1}^{N} w_i \cdot \log(p(y^*|a_i, x))$ represents the log-likelihood of the validation data, while the second term, $\sum_{i=1}^{N} w_i \log(w_i)$, acts as an entropy-based regularizer, discouraging overconfidence in a single prompt.

During inference, model uncertainty was estimated from the weighted entropy of the ensemble of prompts, representing the LLM's confidence in its predictions.

We performed linear optimization of prompt weights for each organ-finding label. We utilized a fixed subset of 50 samples from the manually annotated dataset. We conducted this process independently for each label, resulting in a distinct, fixed weight combination tailored to each organ-finding label.

**Learnable Weights Using MLP:** The uniform and linearly optimized weighting methods assume fixed or simple relationships between prompts, limiting their ability to capture complex interactions. To address this, we introduce a Multi-Layer Perceptron (MLP) for adaptive prompt weight optimization. The MLP dynamically adjusts weights based on prompt behavior across the dataset, enabling more accurate weighting in scenarios with intricate dependencies between prompts. The forward pass for the model is defined as follows:

$$w = softmax(\text{MLP}(h))$$

where $h$ represents the fixed prompt embeddings, and $w$ are the normalized weights.

The training objective combines the binary cross-entropy loss between the ground truth labels $y_{gt}$ and the weighted prompt probabilities $\hat{y}$, with an entropy regularization term to encourage well-calibrated uncertainty:

$$\mathcal{L} = BCE(\hat{y}, y_{gt}) - \lambda \cdot \sum_i w_i \log(w_i)$$

where λ is a hyperparameter that controls the strength of the entropy regularization term.

The MLP is trained using the Adam optimizer, with gradient clipping applied to ensure stable updates. During inference, the final predicted probabilities for each label (yes/no) are the weighted average of the individual prompt probabilities:

$$\hat{y} = \sum_i w_i p_i$$

where $w_i$ are the optimized weights, and $p_i$ are the probabilities associated with each prompt. The final decision is the label with the highest probability. This adaptive approach offers a distinct advantage over simpler methods by dynamically adjusting the prompt weights, potentially improving performance and yielding better-calibrated uncertainty estimates.

We leveraged the automatically labelled cases for the 8 selected labels to train the MLP model. We used 60% of the automatically labelled dataset, comprising 5794 cases, for training, 20% for validation and 20% for internal test of the weights. We trained the MLP model across all labels simultaneously, producing a single model applicable to all labels. During inference, we used the trained MLP model to predict prompt weights dynamically for each case and each of the 23 organ-finding label combinations.

## Evaluation Setup

We evaluated our approach on the manually annotated test dataset (462 reports, with 23 organ–finding combinations), excluding the 50 cases used to tune the linear weights. To establish a baseline, we simulated using any single prompt from our set of six, computing performance metrics for each prompt independently and then averaging the results. We then compared various Bayesian aggregation methods with this baseline, both before and after filtering high-uncertainty cases. For each organ–finding label, we measured accuracy, F1, and Cohen's Kappa at two uncertainty thresholds: excluding samples above 0.5 and excluding up to 20% of samples if their uncertainty exceeded 0.5. We also generated uncertainty histograms for each model, separating correct and incorrect predictions, and reported the median uncertainty in each group to assess how well uncertainty levels distinguish between them.

## Results:

**Uncertainty-aware prediction without filtering**

Table 1 shows that aggregating multiple prompts significantly outperforms using a single prompt. The baseline (single-prompt) model has the lowest metrics, particularly F1=0.2699 and Kappa=0.1790. In contrast, the agent-based approach achieves the best overall balance, with F1=0.3967, recall=0.6437, and Kappa=0.3006, suggesting it handles ambiguity effectively. Although the MLP model attains the highest accuracy (0.8605) and precision (0.3772), its recall is lower (0.3977). Both uniform and linear methods also surpass the baseline, with linear showing slightly higher accuracy (0.8454) for similar F1, emphasizing the benefits of weight-optimized prompt ensembles.

**Uncertainty Histograms**

Figures 3 and 4 show the uncertainty histograms for two representative labels from the four methods, with the Agent model providing the clearest separation between correct and incorrect predictions. Additional histograms appear in Appendix D. Table 2 presents the average median uncertainty across all 23 labels, illustrating that a well-calibrated model should exhibit low uncertainty for correct predictions and high uncertainty for incorrect ones. The Agent model yields an average median uncertainty of 0.0 for correct predictions and about 0.5 for incorrect ones, indicating superior calibration compared to the other methods.

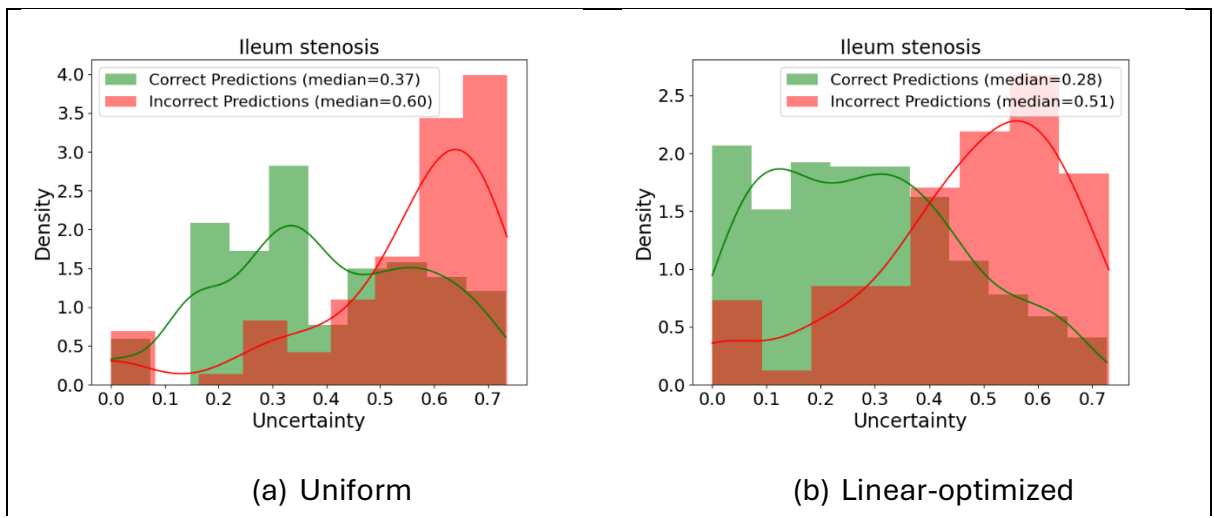

(a) Uniform    (b) Linear-optimized

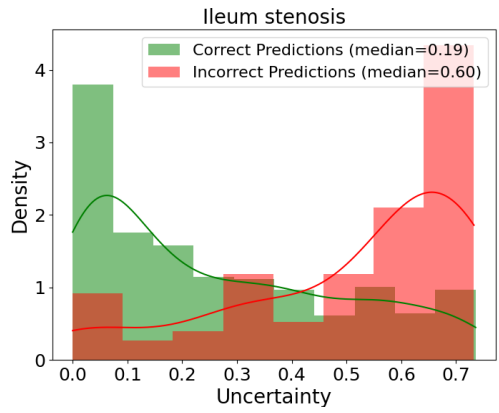
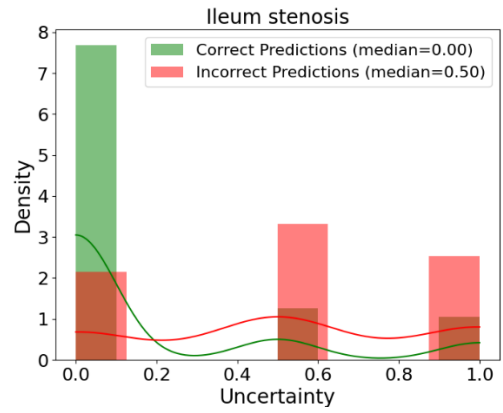

(c) Learnable - MLP    (d) Agent

**Fig. 3:** Uncertainty Histograms for Ileum stenosis Computed by: (a) Uniform Weights, (b) Linear-Optimized Weights, (c) Learnable- MLP Weights, and (d) Agent-based decision.

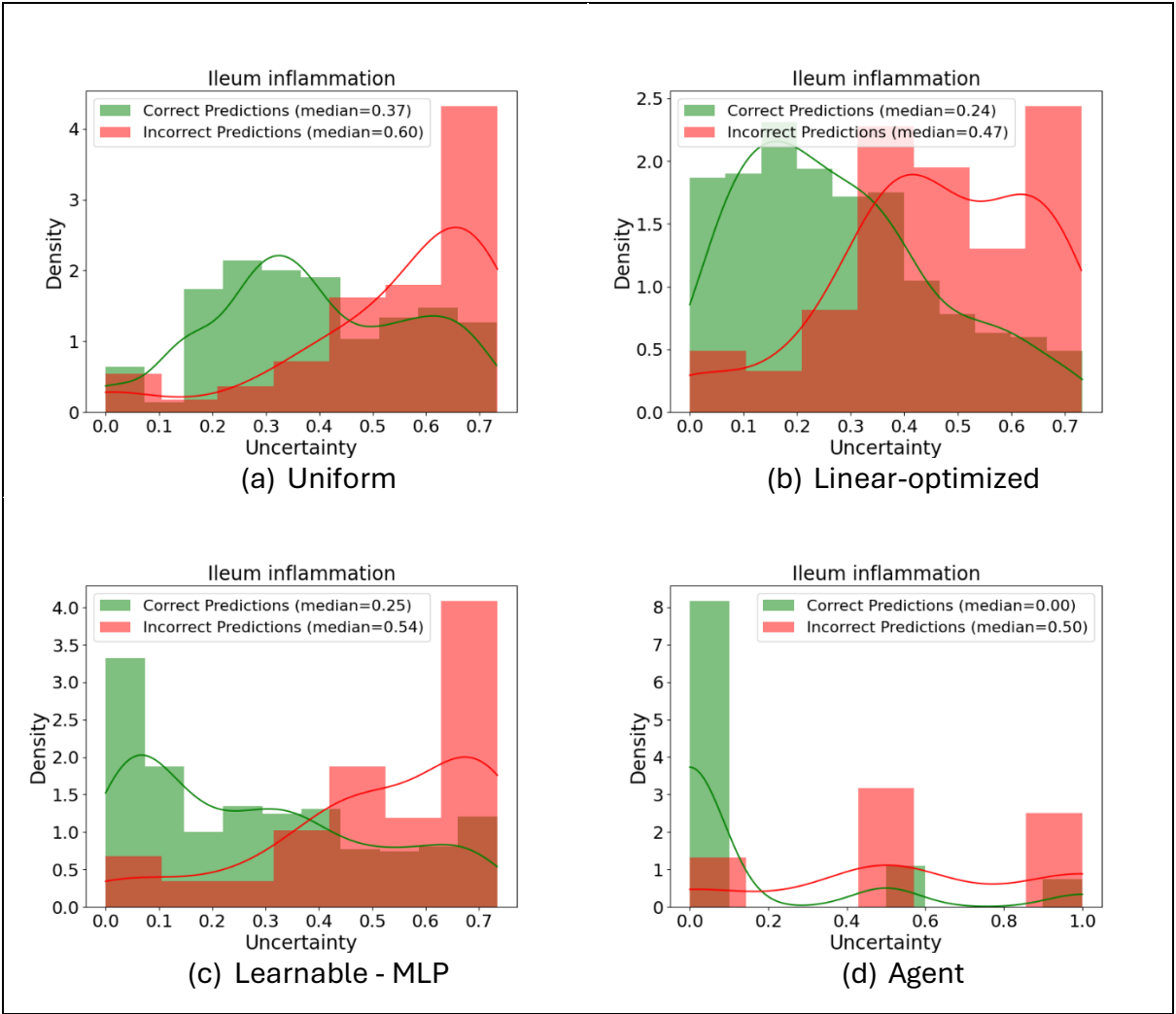

**Fig. 4:** Uncertainty Histograms for Ileum inflammation Computed by: (a) Uniform Weights, (b) Linear-Optimized Weights, (c) Learnable- MLP Weights, and (d) Agent-based decision.

**Uncertainty-aware prediction with filtering**

Tables 3 and 4 present model performance after filtering out cases with uncertainty ≥0.5. Table 3 imposes no cap on excluded cases, while Table 4 limits exclusion to at most 20%. In both scenarios, removing high-uncertainty cases improves accuracy, F1, and Cohen's Kappa. Appendix C provides additional tables summarizing the result per label and model.

With no exclusion cap (Table 3), the MLP model achieves the highest accuracy (92.42%), whereas the Agent approach yields the best F1 (47.87%) and recall (66.14%). Under the 20% cap (Table 4), MLP again leads in accuracy (90.87%), and the Agent method retains the highest F1 (44.94%) and recall (72.95%). These results highlight the Agent method's strong focus on identifying true positives while maintaining robust calibration, even with constrained exclusions.

## Discussion:

In this study, we evaluated prompt-ensemble–based uncertainty awareness for structured data extraction from radiology reports using LLMs. Aggregating predictions from multiple prompts and filtering high-uncertainty cases outperformed single-prompt methods, notably boosting F1 and Cohen's Kappa.

Among the tested approaches, the agent-based method provided the best overall balance in F1, recall, and Cohen's Kappa by integrating multiple prompts and accounting for their consistency. Meanwhile, the MLP model attained the highest precision and accuracy, useful for scenarios demanding fewer false positives, but its lower recall indicates some missed positive cases. This trade-off underscores the importance of selecting models based on task needs.

Additionally, the agent approach demonstrated superior calibration, distinguishing correct and incorrect predictions more effectively than other methods. Entropy-based metrics further confirmed the agent's robust handling of complex prompt outputs, enhancing prediction reliability.

Excluding high-uncertainty cases improved accuracy, F1, and Cohen's Kappa, showing the value of uncertainty-aware filtering for better alignment between model confidence and prediction correctness. Setting class-specific thresholds at the output layer can improve recall for minority classes while preserving precision for majority classes, thus optimizing performance for the target application.

Large language models (LLMs) efficiently automate structured data extraction from radiology reports by converting unstructured text into clinically relevant insights [5-7]. However, ensuring trustworthy application through robust uncertainty quantification remains an open challenge. While Zeng et al. [26] demonstrated the potential of multiple interacting LLM agents with specialized tasks to improve performance in medical applications, their role in quantifying uncertainty for reliable use remains underexplored. This study is the first to introduce a generalizable agent-based approach that quantifies uncertainty and fosters the trustworthy use of LLMs for structured data extraction from radiology reports.

Our study has limitations. First, while the focus was on radiology reports, extending these methods to LLM-based analysis of other medical reports, such as pathology reports or electronic medical records, requires additional validation. Second, optimizing ensemble techniques for highly imbalanced datasets remains a challenge. Future work should explore tailored approaches to address these limitations and investigate the application of uncertainty quantification in broader clinical scenarios.

In conclusion, our findings demonstrate the effectiveness of prompt ensembles-based uncertainty awareness in enhancing LLM performance for structured data extraction in radiology. The agent-based approach emerged as particularly robust, achieving superior results. Incorporating uncertainty quantification not only improves reliability but also facilitates interpretability, paving the way for more impactful and trustable AI applications in healthcare.

**Figure captions:**

**Fig. 1:** (a) Distribution of organ-specific findings and (b) filtered organ-finding combinations with more than 15 positive cases.

**Fig. 2:** Illustration of the proposed Bayesian Prompt Ensemble pipeline for uncertainty-aware predictions. A radiology report and multiple semantically equivalent prompts are fed into an LLM, generating one prediction per prompt. These predictions are then aggregated to yield a final decision and uncertainty estimation. The aggregation function can be either an LLM agent or an entropy-based function that applies uniform weights, learned fixed weights, or weights predicted by a pre-trained MLP model.

**Fig. 3:** Uncertainty Histograms for Ileum stenosis Computed by: (a) Uniform Weights, (b) Linear-Optimized Weights, (c) Learnable- MLP Weights, and (d) Agent-based decision.

**Fig. 4:** Uncertainty Histograms for Ileum inflammation Computed by: (a) Uniform Weights, (b) Linear-Optimized Weights, (c) Learnable- MLP Weights, and (d) Agent-based decision.

**Table 1:** Comparison of average results across all labels for different methods before applying the uncertainty threshold.

| Model | Accuracy | F1 | Precision | Recall | Kappa |
|---|---|---|---|---|---|
| **Baseline** | 0.8243 | 0.2699 | 0.3046 | 0.3097 | 0.179 |
| **Uniform** | 0.832 | 0.3873 | 0.336 | 0.5336 | 0.2979 |
| **Linear** | 0.8454 | 0.3847 | 0.3463 | 0.4878 | 0.2988 |
| **MLP** | **0.8605** | 0.3643 | **0.3772** | 0.3977 | 0.2863 |
| **Agent** | 0.8022 | **0.3967** | 0.3131 | **0.6437** | **0.3006** |

**Table 2:** Comparison of average of the median uncertainty values, across all 23 labels, for correct and incorrect predictions across different methods.

| Method | Median Uncertainty | |
| --- | --- | --- |
| | Correct Predictions | Incorrect Predictions |
| Uniform | 0.4081 | 0.6153 |
| Linear | 0.2668 | 0.4694 |
| MLP | 0.2261 | 0.5717 |
| Agent | **0** | **0.5108** |

**Table 3:** Comparison of average results across all labels for different methods after applying uncertainty threshold ≥0.5.

| Model | Accuracy | F1 | Precision | Recall | Kappa | Excluded Cases |
|---|---|---|---|---|---|---|
| Baseline | 0.8243 | 0.2699 | 0.3046 | 0.3097 | 0.179 | 0 |
| Uniform | 0.9188 | 0.4634 | 0.4458 | 0.5327 | 0.4203 | 200 (43.31%) |
| Linear | 0.897 | 0.4261 | 0.4008 | 0.508 | 0.3702 | 96 (20.97%) |
| MLP | **0.9242** | 0.4032 | **0.4584** | 0.4079 | 0.3643 | 120 (26.07%) |
| Agent | 0.8947 | **0.4787** | 0.3976 | **0.6614** | **0.4258** | 153 (33.16%) |

**Table 4:** Comparison of average results across all labels for different methods after applying uncertainty threshold up to ≥ 0.5, with case exclusions limited to a maximum of 20%.

| Model | Accuracy | F1 | Precision | Recall | Kappa | Excluded Cases |
|---|---|---|---|---|---|---|
| Baseline | 0.8243 | 0.2699 | 0.3046 | 0.3097 | 0.179 | 0 |
| Uniform | 0.8789 | 0.4267 | 0.3873 | 0.5304 | 0.3615 | 92 (19.91%) |
| Linear | 0.8905 | 0.212 | 0.3938 | 0.5079 | 0.3613 | 80 (17.5%) |
| MLP | **0.9087** | 0.3933 | **0.4492** | 0.4004 | 0.345 | 90 (19.5%) |
| Agent | 0.8285 | **0.4494** | 0.3529 | **0.7295** | **0.3683** | 92 (19.91%) |

# Appendix A

**Table S1:** HSMP-BERT performances for the 23 selected labels on the test-set. The selected labels for the MLP training (Kappa>0.7) are highlighted.

| Label | Accuracy | Precision | Recall | F1 | Kappa | Roc-AUC |
|---|---|---|---|---|---|---|
| Cecum inflammation | 0.9310 | 0.7730 | 0.6573 | 0.6959 | 0.3950 | 0.6573 |
| Cecum wall enhancement | 0.9483 | 0.6865 | 0.6865 | 0.6865 | 0.3730 | 0.6865 |
| Cecum wall thickness | 0.9569 | 0.8240 | 0.8611 | 0.8413 | 0.6827 | 0.8611 |
| Colon inflammation | 0.9052 | 0.8228 | 0.7739 | 0.7952 | 0.5910 | 0.7739 |
| Colon mesenteric edema or fat stranding | 0.9569 | 0.4784 | 0.5000 | 0.4890 | 0.0000 | 0.5000 |
| Colon wall enhancement | 0.9310 | 0.8061 | 0.6906 | 0.7432 | 0.6056 | 0.6906 |
| Colon wall thickness | 0.9310 | 0.8394 | 0.8134 | 0.8469 | 0.6945 | 0.8134 |
| **Ileum comb sign** | 0.9483 | 0.9135 | 0.8723 | 0.8913 | 0.7826 | 0.8723 |
| Ileum fistula | 0.9397 | 0.7808 | 0.9146 | 0.8318 | 0.6639 | 0.9146 |
| **Ileum inflammation** | 0.8793 | 0.8745 | 0.8787 | 0.8763 | 0.7527 | 0.8787 |
| Ileum mesenteric edema or fat stranding | 0.8966 | 0.8667 | 0.6513 | 0.6970 | 0.6113 | 0.6513 |
| **Ileum pre-stenotic dilation** | 0.9569 | 0.9212 | 0.9366 | 0.9286 | 0.8573 | 0.9366 |
| Ileum reduced motility | 0.9828 | 0.9912 | 0.7500 | 0.8290 | 0.6588 | 0.7500 |
| **Ileum stenosis** | 0.9224 | 0.8884 | 0.9138 | 0.9008 | 0.8000 | 0.9138 |
| **Ileum wall enhancement** | 0.9569 | 0.9514 | 0.9559 | 0.9536 | 0.9072 | 0.9559 |
| **Ileum wall thickness** | 0.9397 | 0.9417 | 0.9444 | 0.9396 | 0.8797 | 0.9444 |
| Rectum inflammation | 0.9397 | 0.8482 | 0.6620 | 0.7148 | 0.4345 | 0.6620 |
| Rectum wall enhancement | 0.9714 | 0.9867 | 0.7500 | 0.8261 | 0.6548 | 0.7500 |
| **Rectum wall thickness** | 0.9828 | 0.9910 | 0.8571 | 0.9121 | 0.8245 | 0.8571 |
| **Sigmoid comb sign** | 0.9914 | 0.9956 | 0.8333 | 0.8978 | 0.7958 | 0.8333 |
| Sigmoid inflammation | 0.8793 | 0.7484 | 0.6938 | 0.7157 | 0.4330 | 0.6938 |
| Sigmoid wall enhancement | 0.9310 | 0.7730 | 0.6573 | 0.6959 | 0.3950 | 0.6573 |
| Sigmoid wall thickness | 0.9483 | 0.9144 | 0.7869 | 0.8358 | 0.6729 | 0.7869 |

## Appendix B

> **The different prompts used to query the LLM for the data in the radiology report.**

**Does the following radiology report indicate that the patient has {*label*}? Here is the report: {*report*}. Answer shortly, format your answer as a JSON file using the following schema: {"Answer": "Yes/No", "Explanation": "str"}. Important: Only return a single piece of valid JSON text. For example: {"Answer": "Yes", "Explanation": "The report mentions that the patient has a severe case of the disease."}**

**Does the radiology report below suggest that the patient is suffering from {*label*}? Here is the report: {*report*}. Please provide a concise answer, formatted as a JSON object using the following schema: {"Answer": "Yes/No", "Explanation": "str"}. Important: Ensure you return a single valid JSON object. For instance: {"Answer": "Yes", "Explanation": "The report indicates that the patient has severe symptoms of the condition."}**

**Can the presence of {*label*} be confirmed from the following radiology report? Here is the report: {*report*}.  Provide a brief response, formatted as a JSON object according to the following schema: {"Answer": "Yes/No", "Explanation": "str"}. Important: Only return a single valid JSON object.**

**Does the patient appear to have {*label*} based on the analysis of the radiology report provided? Here is the report: {*report*}. Please answer succinctly, formatted as a JSON object in the following schema: {"Answer": "Yes/No", "Explanation": "str"}. Important: Return only a single valid JSON object. For example: {"Answer": "Yes", "Explanation": "The report clearly states that the patient has a significant case of the condition."}**

**Considering the radiology report presented, is there an indication that the patient has {*label*}? Here is the report: {*report*}. Provide a brief answer, formatted as a JSON object according to the schema: {"Answer": "Yes/No", "Explanation": "str"}. Important: Make sure to return a single valid JSON object. For instance: {"Answer": "Yes", "Explanation": "The report notes that the patient shows clear signs of the condition."}**

**Is it possible to conclude from the following radiology report that the patient has {*label*}? Here is the report: {*report*}. Please provide a concise response, formatted as a JSON object using the schema: {"Answer": "Yes/No", "Explanation": "str"}. Important: Ensure only a single valid JSON object is returned. For instance: {"Answer": "Yes", "Explanation": "The report suggests that the patient has a pronounced case of the condition."}**

# Appendix C

Comparison Results for Different Uncertainty Thresholding.

**Table S2:** Comparison of different methods before filtering using uncertainty threshold.

| Label | prevalence | F1 Score [%] | | | | | Accuracy [%] | | | | | Kappa [%] | | | | |
|---|---|---|---|---|---|---|---|---|---|---|---|---|---|---|---|---|
| | | Baseline | Uniform | Linear | MLP | Agent | Baseline | Uniform | Linear | MLP | Agent | Baseline | Uniform | Linear | MLP | Agent |
| Cecum inflammation | 8.22% | 25.415 | 28.148 | 29.508 | 28.829 | 33.083 | 83.117 | 79.004 | 81.385 | 82.900 | 80.736 | 0.169 | 0.185 | 0.205 | 0.202 | 0.242 |
| Cecum wall enhancement | 4.54% | 9.525 | 12.000 | 12.048 | 11.429 | 11.475 | 84.776 | 80.952 | 84.199 | 86.580 | 76.623 | 0.034 | 0.052 | 0.056 | 0.054 | 0.043 |
| Cecum wall thickness | 6.49% | 24.929 | 32.759 | 31.858 | 32.911 | 27.419 | 85.426 | 83.117 | 83.333 | 88.528 | 80.519 | 0.181 | 0.256 | 0.247 | 0.270 | 0.195 |
| Colon inflammation | 15.15% | 37.020 | 44.954 | 44.118 | 42.786 | 43.478 | 77.633 | 74.026 | 75.325 | 75.108 | 69.048 | 0.239 | 0.307 | 0.302 | 0.287 | 0.276 |
| Colon mesenteric edema or fat stranding | 3.24% | 4.893 | 8.889 | 10.000 | 13.793 | 18.667 | 91.522 | 91.126 | 92.208 | 94.589 | 86.797 | 0.016 | 0.048 | 0.062 | 0.110 | 0.142 |
| Colon wall enhancement | 8.22% | 16.365 | 23.913 | 24.390 | 20.290 | 28.788 | 84.632 | 84.848 | 86.580 | 88.095 | 79.654 | 0.091 | 0.158 | 0.171 | 0.139 | 0.193 |
| Colon wall thickness | 12.77% | 29.524 | 41.905 | 39.604 | 38.636 | 45.802 | 84.127 | 86.797 | 86.797 | 88.312 | 84.632 | 0.214 | 0.346 | 0.324 | 0.330 | 0.370 |
| Ileum comb sign | 14.50% | 28.299 | 41.808 | 39.216 | 31.746 | 42.927 | 77.814 | 77.706 | 79.870 | 81.385 | 74.675 | 0.165 | 0.290 | 0.274 | 0.210 | 0.291 |
| Ileum fistula | 6.27% | 33.782 | 48.421 | 51.282 | 54.286 | 43.137 | 87.879 | 89.394 | 91.775 | 93.074 | 87.446 | 0.282 | 0.435 | 0.471 | 0.507 | 0.375 |
| Ileum inflammation | 41.77% | 60.544 | 86.162 | 84.182 | 84.615 | 86.978 | 73.846 | 88.528 | 87.229 | 87.879 | 88.528 | 0.430 | 0.764 | 0.735 | 0.747 | 0.768 |
| Ileum mesenteric edema or fat stranding | 11.90% | 17.608 | 34.000 | 26.966 | 26.506 | 36.842 | 84.488 | 85.714 | 85.931 | 86.797 | 84.416 | 0.105 | 0.261 | 0.197 | 0.201 | 0.280 |
| Ileum pre-stenotic dilation | 16.01% | 27.070 | 40.816 | 42.105 | 40.698 | 39.583 | 73.052 | 68.615 | 71.429 | 77.922 | 62.338 | 0.121 | 0.238 | 0.261 | 0.275 | 0.207 |
| Ileum reduced motility | 2.81% | 17.169 | 26.316 | 33.333 | 28.000 | 20.000 | 87.518 | 87.879 | 92.208 | 92.208 | 82.684 | 0.135 | 0.227 | 0.304 | 0.249 | 0.159 |
| Ileum stenosis | 22.07% | 39.536 | 56.585 | 53.608 | 53.631 | 64.655 | 78.571 | 80.736 | 80.519 | 82.035 | 82.251 | 0.278 | 0.442 | 0.413 | 0.428 | 0.530 |
| Ileum wall enhancement | 31.88% | 40.055 | 65.815 | 70.414 | 50.000 | 67.630 | 68.872 | 76.790 | 78.308 | 73.102 | 75.705 | 0.212 | 0.483 | 0.537 | 0.325 | 0.489 |
| Ileum wall thickness | 42.85% | 56.885 | 79.656 | 77.551 | 73.457 | 88.384 | 72.367 | 84.632 | 83.333 | 81.385 | 90.043 | 0.399 | 0.677 | 0.648 | 0.602 | 0.797 |
| Rectum inflammation | 7.14% | 21.380 | 27.273 | 28.261 | 31.461 | 26.087 | 85.281 | 82.684 | 85.714 | 86.797 | 77.922 | 0.140 | 0.192 | 0.210 | 0.247 | 0.171 |
| Rectum wall enhancement | 3.67% | 8.962 | 20.690 | 18.868 | 11.765 | 20.690 | 89.935 | 90.043 | 90.693 | 93.506 | 85.065 | 0.051 | 0.163 | 0.146 | 0.084 | 0.157 |
| Rectum wall thickness | 5.84% | 25.029 | 33.803 | 29.730 | 37.736 | 36.893 | 89.610 | 89.827 | 88.745 | 92.857 | 85.931 | 0.200 | 0.286 | 0.241 | 0.339 | 0.309 |
| Sigmoid comb sign | 3.46% | 16.065 | 21.918 | 25.000 | 18.868 | 18.367 | 88.348 | 87.662 | 90.909 | 90.693 | 82.684 | 0.117 | 0.175 | 0.211 | 0.147 | 0.133 |
| Sigmoid inflammation | 12.55% | 29.698 | 40.000 | 37.255 | 39.362 | 36.015 | 75.433 | 71.429 | 72.294 | 75.325 | 63.853 | 0.164 | 0.264 | 0.235 | 0.266 | 0.205 |
| Sigmoid wall enhancement | 7.14% | 17.963 | 27.368 | 23.377 | 23.729 | 33.824 | 84.957 | 85.065 | 87.229 | 90.260 | 80.519 | 0.112 | 0.199 | 0.166 | 0.186 | 0.258 |
| Sigmoid wall thickness | 11.03% | 33.069 | 47.788 | 52.252 | 43.373 | 41.791 | 86.869 | 87.229 | 88.528 | 89.827 | 83.117 | 0.264 | 0.406 | 0.458 | 0.381 | 0.326 |

**Table S3:** Comparison of different methods after filtering using uncertainty threshold < 0.5.

| Label | F1 Score | | | | | Accuracy | | | | | Kappa | | | | | Cases | | | | |
|---|---|---|---|---|---|---|---|---|---|---|---|---|---|---|---|---|---|---|---|---|
| | Baseline | Uniform | Linear | MLP | Agent | Baseline | Uniform | Linear | MLP | Agent | Baseline | Uniform | Linear | MLP | Agent | Baseline | Uniform | Linear | MLP | Agent |
| Cecum inflammation | 25.415 | 37.209 | 28.571 | 35.294 | 33.333 | 83.117 | 90.217 | 88.439 | 93.060 | 88.991 | 0.169 | 0.324 | 0.235 | 0.317 | 0.291 | 462 | 276 | 346 | 317 | 327 |
| Cecum wall enhancement | 9.525 | 14.286 | 15.385 | 7.143 | 13.636 | 84.776 | 90.123 | 87.978 | 92.145 | 88.013 | 0.034 | 0.104 | 0.101 | 0.030 | 0.089 | 462 | 243 | 366 | 331 | 317 |
| Cecum wall thickness | 24.929 | 59.259 | 47.619 | 41.667 | 40 | 85.426 | 95.971 | 93.189 | 95.796 | 91.429 | 0.181 | 0.573 | 0.442 | 0.395 | 0.363 | 462 | 273 | 323 | 333 | 315 |
| Colon inflammation | 37.020 | 54.237 | 49.645 | 51.724 | 51.701 | 77.633 | 81.315 | 80.759 | 83.673 | 77.389 | 0.239 | 0.437 | 0.391 | 0.423 | 0.399 | 462 | 289 | 369 | 343 | 314 |
| Colon mesenteric edema or fat stranding | 4.893 | 20 | 18.182 | 26.667 | 28.571 | 91.522 | 97.403 | 95.522 | 97.158 | 95.268 | 0.016 | 0.187 | 0.159 | 0.255 | 0.264 | 462 | 308 | 402 | 387 | 317 |
| Colon wall enhancement | 16.365 | 14.815 | 22.222 | 16.216 | 17.021 | 84.632 | 91.513 | 89.231 | 91.507 | 87.171 | 0.091 | 0.104 | 0.165 | 0.122 | 0.110 | 462 | 271 | 390 | 365 | 304 |
| Colon wall thickness | 29.524 | 53.333 | 46.667 | 39.024 | 58.182 | 84.127 | 94.928 | 91.557 | 93.075 | 92.557 | 0.214 | 0.507 | 0.423 | 0.366 | 0.541 | 462 | 276 | 379 | 361 | 309 |
| Ileum comb sign | 28.299 | 50 | 33.803 | 44.898 | 45.070 | 77.814 | 91.111 | 86.610 | 91.589 | 84.337 | 0.165 | 0.451 | 0.264 | 0.404 | 0.369 | 462 | 180 | 351 | 321 | 249 |
| Ileum fistula | 33.782 | 72.727 | 52.381 | 68.966 | 71.795 | 87.879 | 96.774 | 95.025 | 97.656 | 96.440 | 0.282 | 0.710 | 0.500 | 0.678 | 0.700 | 462 | 279 | 402 | 384 | 309 |
| Ileum inflammation | 60.544 | 93.103 | 87.719 | 89.270 | 96.350 | 73.846 | 94.595 | 90.814 | 92.795 | 97.093 | 0.430 | 0.887 | 0.804 | 0.839 | 0.939 | 462 | 296 | 381 | 347 | 344 |
| Ileum mesenteric edema or fat stranding | 17.608 | 16.667 | 27.451 | 0 | 33.333 | 84.488 | 93.174 | 90.339 | 92.800 | 92.429 | 0.105 | 0.140 | 0.229 | -0.010 | 0.294 | 462 | 293 | 383 | 375 | 317 |
| Ileum pre-stenotic dilation | 27.070 | 60.606 | 55.285 | 55.072 | 47.788 | 73.052 | 84.049 | 81.544 | 87.398 | 75.918 | 0.121 | 0.512 | 0.448 | 0.480 | 0.355 | 462 | 163 | 298 | 246 | 245 |
| Ileum reduced motility | 17.169 | 40 | 32 | 33.333 | 44.444 | 87.518 | 96.203 | 95.550 | 97.640 | 94.604 | 0.135 | 0.385 | 0.303 | 0.323 | 0.421 | 462 | 237 | 382 | 339 | 278 |
| Ileum stenosis | 39.536 | 69.231 | 62.069 | 57.971 | 79.245 | 78.571 | 91.367 | 87.912 | 91.265 | 92.994 | 0.278 | 0.642 | 0.549 | 0.535 | 0.751 | 462 | 278 | 364 | 332 | 314 |
| Ileum wall enhancement | 40.055 | 69.565 | 73.548 | 57.143 | 67.114 | 68.872 | 85.641 | 83.730 | 86.121 | 82.182 | 0.212 | 0.603 | 0.620 | 0.489 | 0.556 | 461 | 195 | 252 | 281 | 275 |
| Ileum wall thickness | 56.885 | 91.971 | 83.810 | 81.633 | 95.575 | 72.367 | 95.492 | 90.368 | 91.641 | 96.960 | 0.399 | 0.888 | 0.771 | 0.764 | 0.933 | 462 | 244 | 353 | 323 | 329 |
| Rectum inflammation | 21.380 | 36.842 | 32.877 | 38.095 | 36.923 | 85.281 | 91.549 | 88.305 | 92.614 | 87.538 | 0.140 | 0.325 | 0.269 | 0.342 | 0.310 | 462 | 284 | 419 | 352 | 329 |
| Rectum wall enhancement | 8.962 | 0 | 20.690 | 0 | 28.571 | 89.935 | 94.872 | 94.148 | 96.296 | 93.528 | 0.051 | -0.026 | 0.178 | -0.017 | 0.261 | 462 | 273 | 393 | 378 | 309 |
| Rectum wall thickness | 25.029 | 58.824 | 40 | 38.095 | 50 | 89.610 | 97.569 | 94.167 | 96.359 | 93.902 | 0.200 | 0.576 | 0.370 | 0.364 | 0.471 | 462 | 288 | 360 | 357 | 328 |
| Sigmoid comb sign | 16.065 | 18.750 | 20.513 | 23.077 | 32.432 | 88.348 | 90.780 | 92.512 | 94.565 | 92.447 | 0.117 | 0.146 | 0.174 | 0.204 | 0.291 | 462 | 282 | 414 | 368 | 331 |
| Sigmoid inflammation | 29.698 | 42.222 | 43.548 | 43.478 | 45.763 | 75.433 | 78.151 | 79.290 | 82.253 | 77.224 | 0.164 | 0.319 | 0.335 | 0.346 | 0.350 | 462 | 238 | 338 | 293 | 281 |
| Sigmoid wall enhancement | 17.963 | 31.579 | 35.294 | 31.250 | 32 | 84.957 | 94.961 | 93.732 | 93.872 | 89.032 | 0.112 | 0.290 | 0.320 | 0.281 | 0.269 | 462 | 258 | 351 | 359 | 310 |
| Sigmoid wall thickness | 33.069 | 60.606 | 50.847 | 47.368 | 52.174 | 86.869 | 95.652 | 92.388 | 94.490 | 90.571 | 0.264 | 0.584 | 0.468 | 0.450 | 0.470 | 462 | 299 | 381 | 363 | 350 |

**Table S4:** Comparison of different methods after filtering up to 20% of the data, using uncertainty threshold < 0.5.

| Label | F1 Score | | | | | Accuracy | | | | | Kappa | | | | | Cases | | | | |
|---|---|---|---|---|---|---|---|---|---|---|---|---|---|---|---|---|---|---|---|---|
| | Baseline | Uniform | Linear | MLP | Agent | Baseline | Uniform | Linear | MLP | Agent | Baseline | Uniform | Linear | MLP | Agent | Baseline | Uniform | Linear | MLP | Agent |
| Cecum inflammation | 25.415 | 31.325 | 33.333 | 35.088 | 36.364 | 83.117 | 84.595 | 88.108 | 90 | 82.973 | 0.169 | 0.245 | 0.277 | 0.299 | 0.294 | 462 | 370 | 370 | 370 | 370 |
| Cecum wall enhancement | 9.525 | 15.625 | 15.385 | 15 | 14.894 | 84.776 | 85.405 | 88.108 | 90.811 | 78.378 | 0.034 | 0.095 | 0.102 | 0.103 | 0.083 | 462 | 370 | 370 | 370 | 370 |
| Cecum wall thickness | 24.929 | 38.710 | 42.623 | 35.897 | 30.928 | 85.426 | 89.730 | 90.541 | 93.243 | 81.892 | 0.181 | 0.342 | 0.379 | 0.324 | 0.244 | 462 | 370 | 370 | 370 | 370 |
| Colon inflammation | 37.020 | 49.080 | 49.296 | 50 | 50.256 | 77.633 | 77.568 | 80.541 | 81.622 | 73.784 | 0.239 | 0.363 | 0.387 | 0.394 | 0.362 | 462 | 370 | 370 | 370 | 370 |
| Colon mesenteric edema or fat stranding | 4.893 | 16.667 | 18.182 | 26.667 | 17.857 | 91.522 | 94.595 | 95.522 | 97.158 | 87.568 | 0.016 | 0.140 | 0.159 | 0.255 | 0.137 | 462 | 370 | 402 | 387 | 370 |
| Colon wall enhancement | 16.365 | 28.070 | 22.222 | 16.216 | 30.189 | 84.632 | 88.919 | 89.231 | 91.622 | 80 | 0.091 | 0.221 | 0.165 | 0.122 | 0.212 | 462 | 370 | 390 | 370 | 370 |
| Colon wall thickness | 29.524 | 47.761 | 46.667 | 40.909 | 54.902 | 84.127 | 90.541 | 91.557 | 92.973 | 87.568 | 0.214 | 0.427 | 0.423 | 0.383 | 0.481 | 462 | 370 | 379 | 370 | 370 |
| Ileum comb sign | 28.299 | 41.584 | 35.443 | 36.620 | 47.853 | 77.814 | 84.054 | 86.216 | 87.838 | 77.027 | 0.165 | 0.329 | 0.277 | 0.300 | 0.358 | 462 | 370 | 370 | 370 | 370 |
| Ileum fistula | 33.782 | 65.385 | 52.381 | 68.966 | 53.012 | 87.879 | 95.135 | 95.025 | 97.656 | 89.459 | 0.282 | 0.629 | 0.500 | 0.678 | 0.481 | 462 | 370 | 402 | 384 | 370 |
| Ileum inflammation | 60.544 | 90.604 | 87.719 | 88.281 | 92.652 | 73.846 | 92.432 | 90.814 | 91.892 | 93.784 | 0.430 | 0.843 | 0.804 | 0.821 | 0.873 | 462 | 370 | 381 | 370 | 370 |
| Ileum mesenteric edema or fat stranding | 17.608 | 29.787 | 27.451 | 0 | 45.977 | 84.488 | 91.081 | 90.339 | 92.800 | 87.297 | 0.105 | 0.252 | 0.229 | -0.010 | 0.389 | 462 | 370 | 383 | 375 | 370 |
| Ileum pre-stenotic dilation | 27.070 | 46.154 | 46.707 | 46.667 | 43.902 | 73.052 | 73.514 | 75.946 | 82.703 | 62.703 | 0.121 | 0.314 | 0.339 | 0.367 | 0.243 | 462 | 370 | 370 | 370 | 370 |
| Ileum reduced motility | 17.169 | 31.579 | 32 | 37.500 | 21.687 | 87.518 | 92.973 | 95.550 | 97.297 | 82.432 | 0.135 | 0.289 | 0.303 | 0.363 | 0.174 | 462 | 370 | 382 | 370 | 370 |
| Ileum stenosis | 39.536 | 63.704 | 61.667 | 60.606 | 76.471 | 78.571 | 86.757 | 87.568 | 89.459 | 89.189 | 0.278 | 0.556 | 0.543 | 0.549 | 0.696 | 462 | 370 | 370 | 370 | 370 |
| Ileum wall enhancement | 40.055 | 66.964 | 72.797 | 53.988 | 74.126 | 68.872 | 79.946 | 80.759 | 79.675 | 79.946 | 0.212 | 0.527 | 0.583 | 0.412 | 0.585 | 461 | 369 | 369 | 369 | 369 |
| Ileum wall thickness | 56.885 | 86.853 | 83.700 | 79.245 | 93.836 | 72.367 | 91.081 | 90 | 88.108 | 95.135 | 0.399 | 0.802 | 0.766 | 0.713 | 0.898 | 462 | 370 | 370 | 370 | 370 |
| Rectum inflammation | 21.380 | 38.095 | 32.877 | 36 | 29.167 | 85.281 | 89.459 | 88.305 | 91.351 | 81.622 | 0.140 | 0.327 | 0.269 | 0.314 | 0.211 | 462 | 370 | 419 | 370 | 370 |
| Rectum wall enhancement | 8.962 | 7.692 | 20.690 | 0 | 23.188 | 89.935 | 93.514 | 94.148 | 96.296 | 85.676 | 0.051 | 0.045 | 0.178 | -0.017 | 0.191 | 462 | 370 | 393 | 378 | 370 |
| Rectum wall thickness | 25.029 | 42.857 | 41.026 | 36.364 | 44.706 | 89.610 | 93.514 | 93.784 | 96.216 | 87.297 | 0.200 | 0.394 | 0.378 | 0.347 | 0.390 | 462 | 370 | 370 | 370 | 370 |
| Sigmoid comb sign | 16.065 | 18.182 | 20.513 | 23.077 | 23.529 | 88.348 | 90.270 | 92.512 | 94.595 | 85.946 | 0.117 | 0.144 | 0.174 | 0.204 | 0.189 | 462 | 370 | 414 | 370 | 370 |
| Sigmoid inflammation | 29.698 | 39.752 | 43.662 | 40.310 | 41.475 | 75.433 | 73.784 | 78.378 | 79.189 | 65.676 | 0.164 | 0.272 | 0.331 | 0.300 | 0.256 | 462 | 370 | 370 | 370 | 370 |
| Sigmoid wall enhancement | 17.963 | 33.333 | 31.579 | 28.571 | 37.383 | 84.957 | 90.270 | 92.973 | 93.243 | 81.892 | 0.112 | 0.284 | 0.279 | 0.251 | 0.293 | 462 | 370 | 370 | 370 | 370 |
| Sigmoid wall thickness | 33.069 | 51.724 | 50.847 | 48.780 | 49.412 | 86.869 | 92.432 | 92.388 | 94.324 | 88.378 | 0.264 | 0.477 | 0.468 | 0.463 | 0.430 | 462 | 370 | 381 | 370 | 370 |

# Appendix D

Uncertainty histograms.

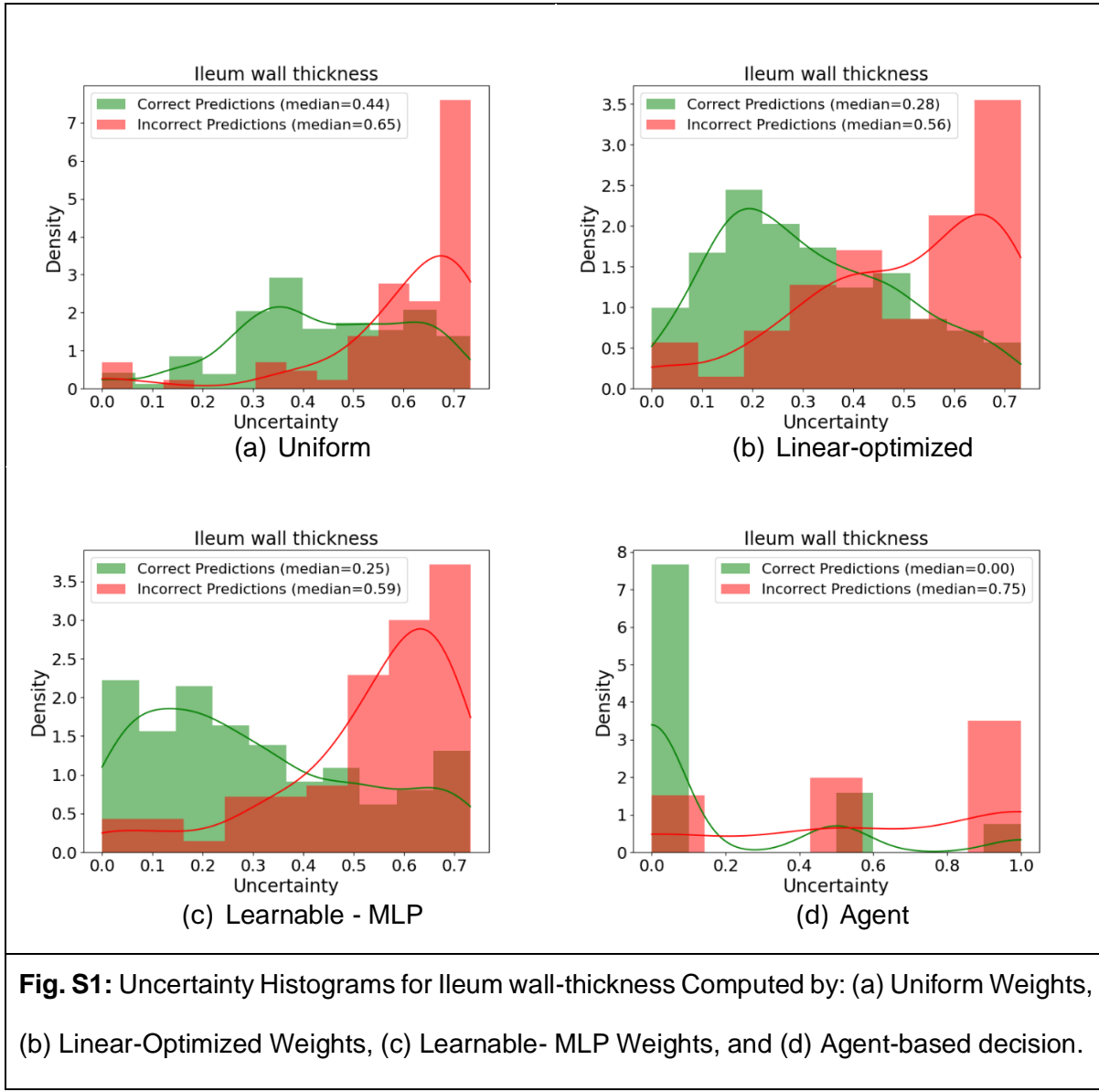

**Fig. S1:** Uncertainty Histograms for Ileum wall-thickness Computed by: (a) Uniform Weights, (b) Linear-Optimized Weights, (c) Learnable- MLP Weights, and (d) Agent-based decision.

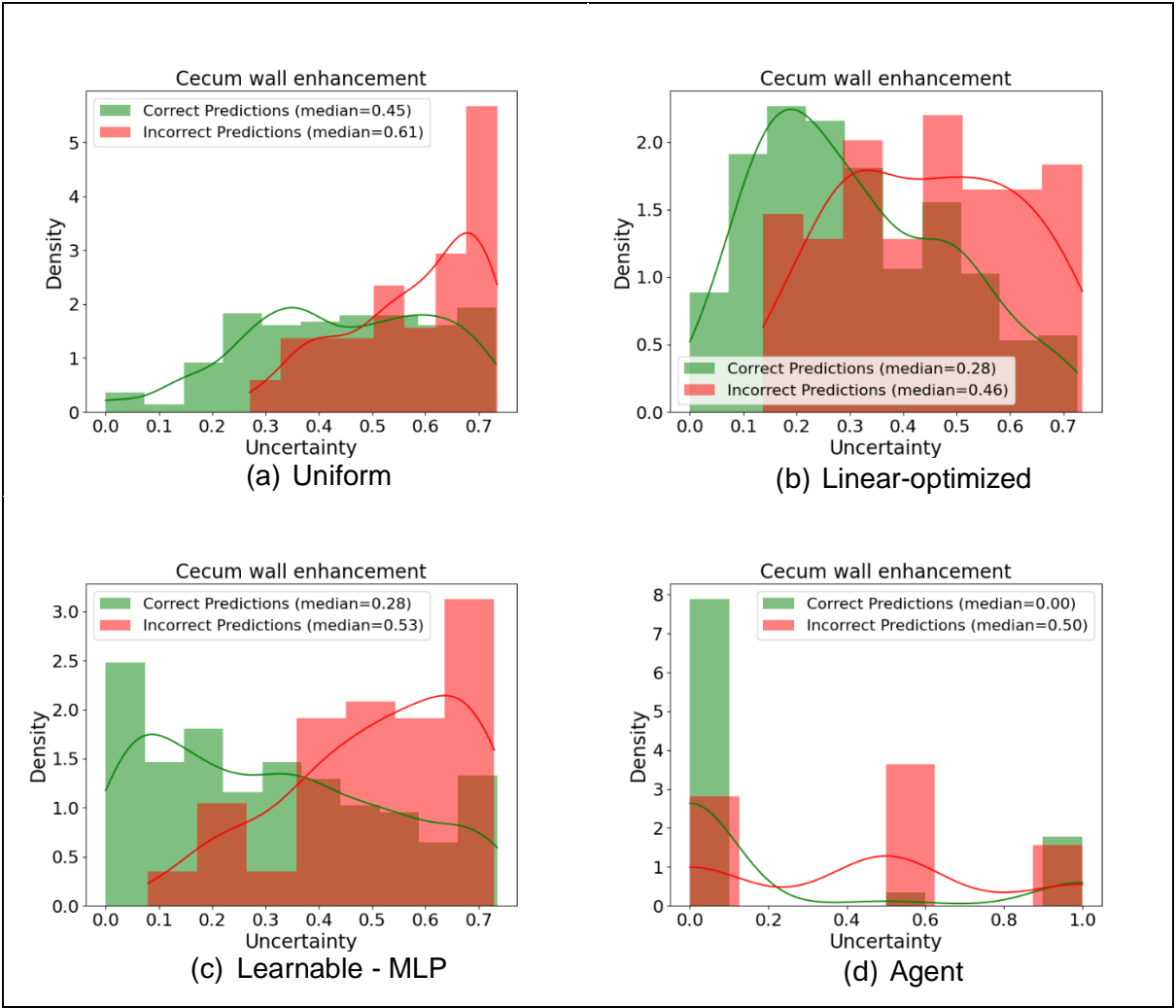

**Fig. S2:** Uncertainty Histograms for Cecum Wall Enhancement computed by: (a) Uniform Weights, (b) Linear-Optimized Weights, (c) Learnable-MLP Weights, and (d) Agent-based decision.

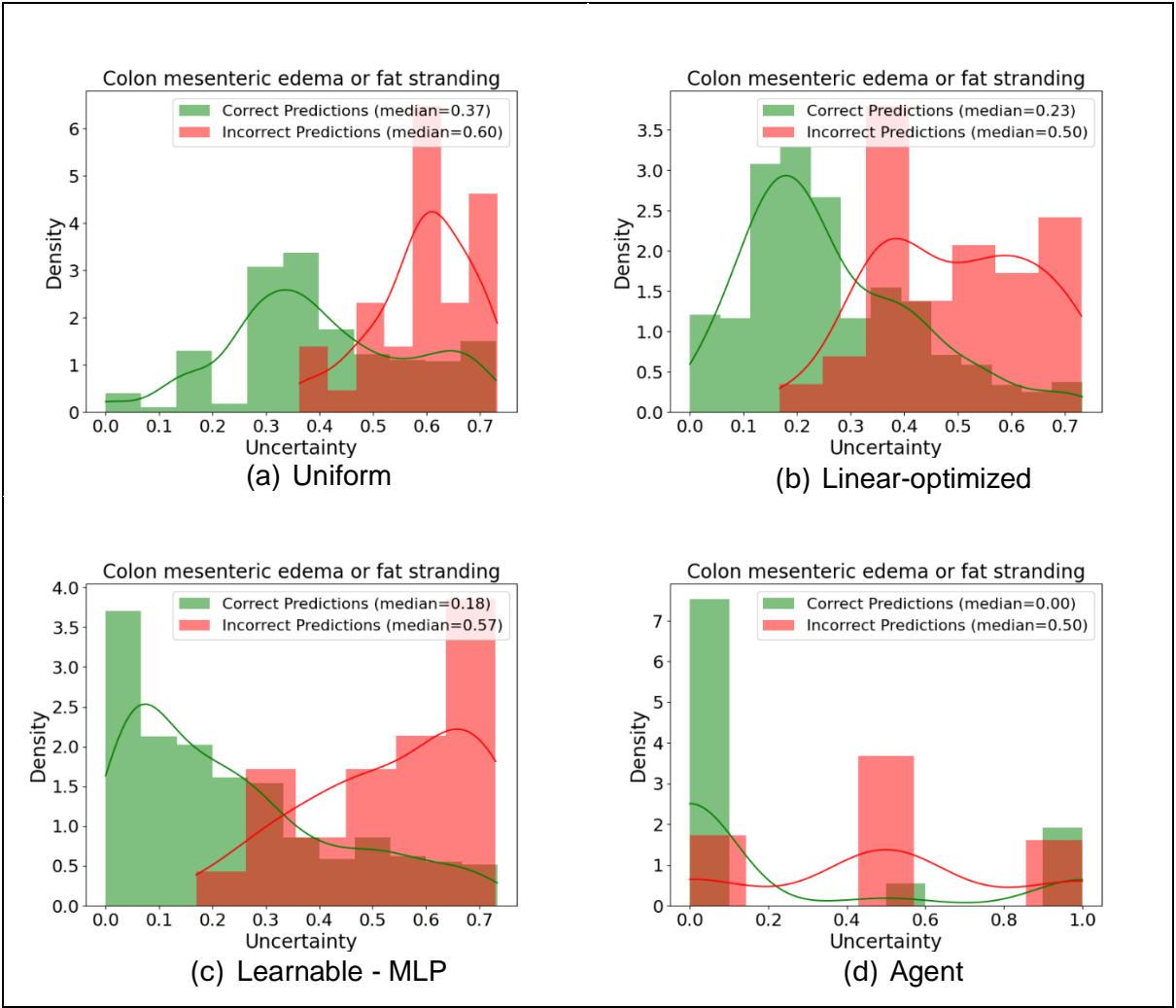

**Fig. S3:** Uncertainty Histograms for Colon Mesenteric Edema or Fat Stranding computed by: (a) Uniform Weights, (b) Linear-Optimized Weights, (c) Learnable-MLP Weights, and (d) Agent-based decision.

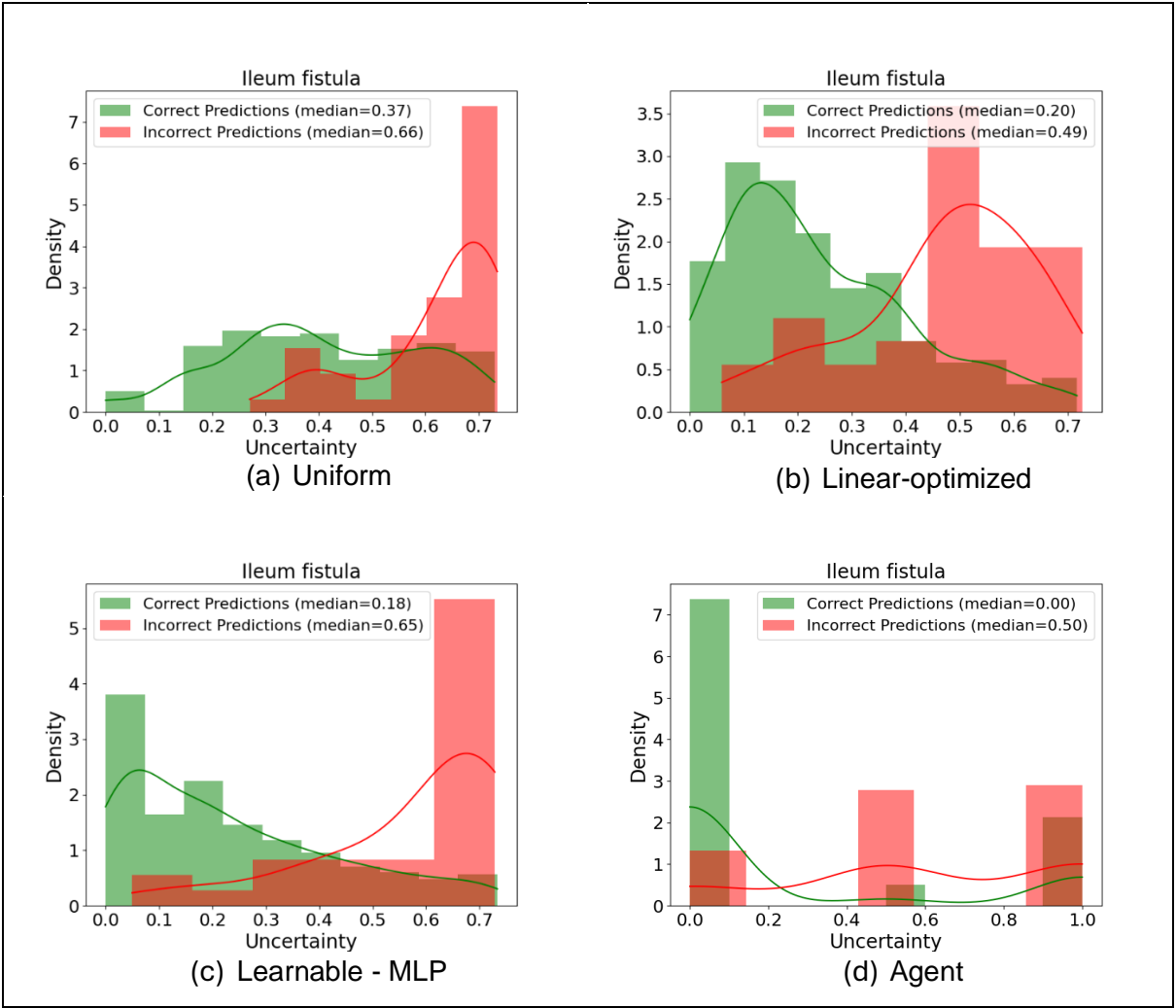

**Fig. S4:** Uncertainty Histograms for Ileum fistula computed by: (a) Uniform Weights, (b) Linear-Optimized Weights, (c) Learnable-MLP Weights, and (d) Agent-based decision.

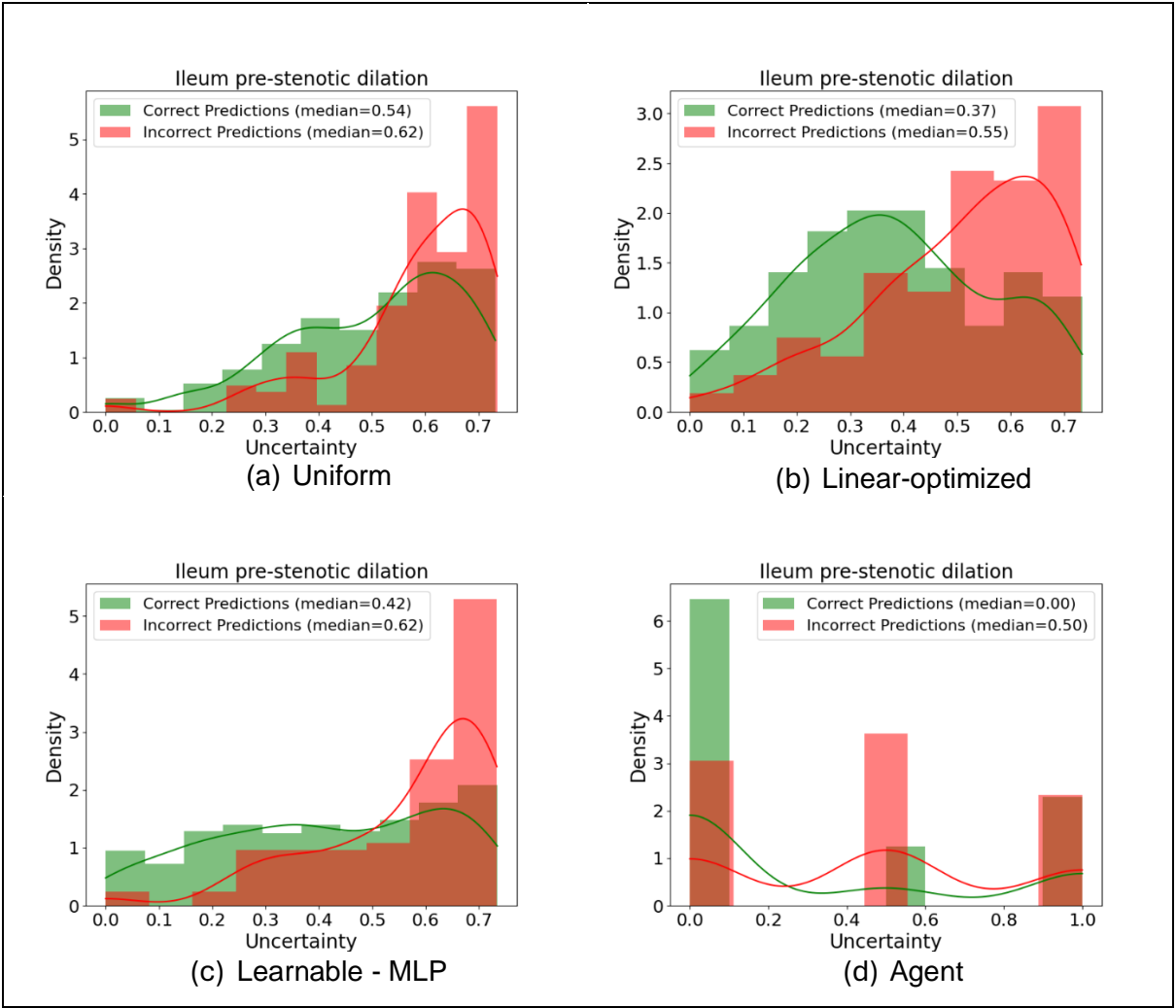

**Fig. S5:** Uncertainty Histograms for Ileum pre-stenotic dilation computed by: (a) Uniform Weights, (b) Linear-Optimized Weights, (c) Learnable-MLP Weights, and (d) Agent-based decision.

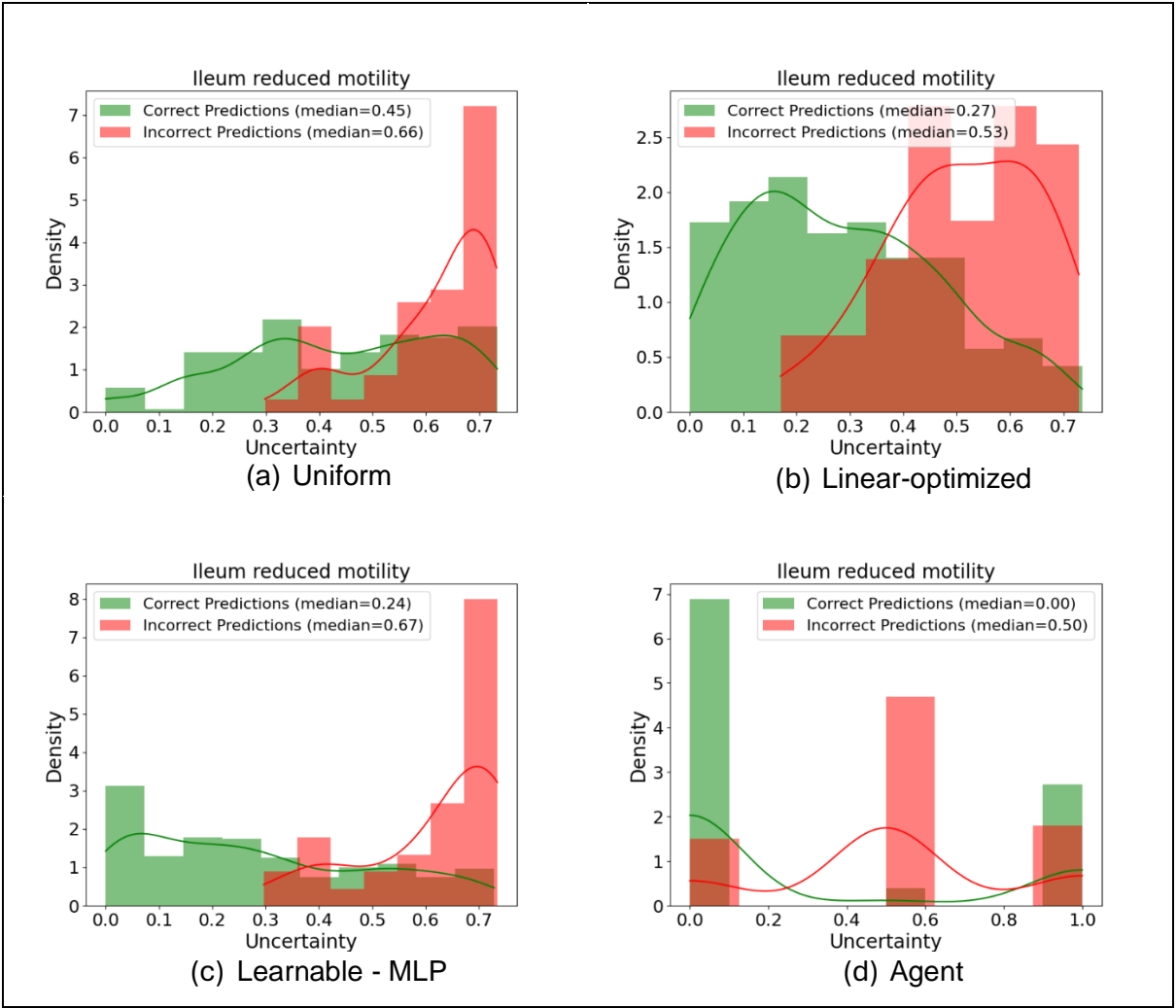

**Fig. S6:** Uncertainty Histograms for Ileum reduced motility computed by: (a) Uniform Weights, (b) Linear-Optimized Weights, (c) Learnable-MLP Weights, and (d) Agent-based decision.

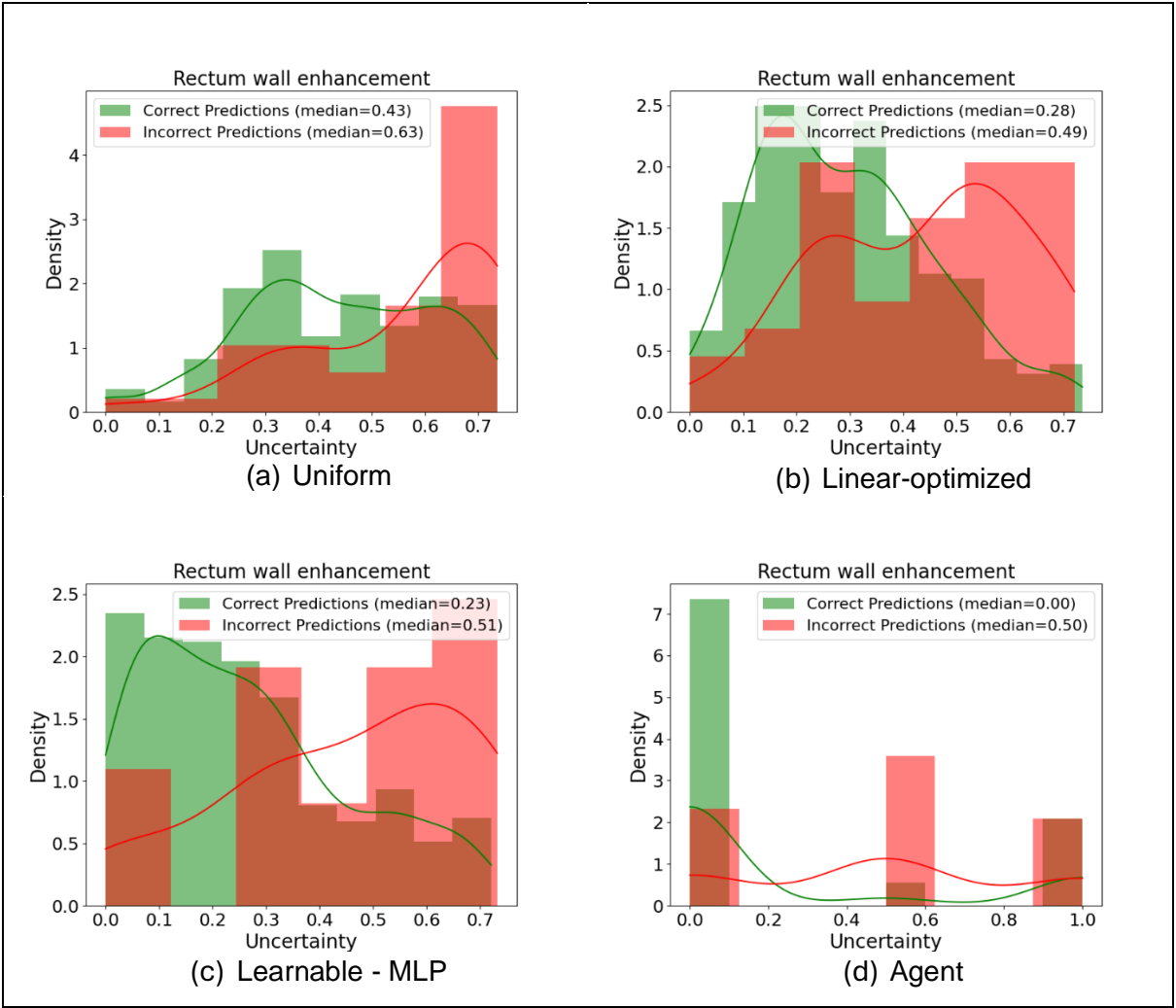

**Fig. S7:** Uncertainty Histograms for Rectum Wall Enhancement computed by: (a) Uniform Weights, (b) Linear-Optimized Weights, (c) Learnable-MLP Weights, and (d) Agent-based decision.

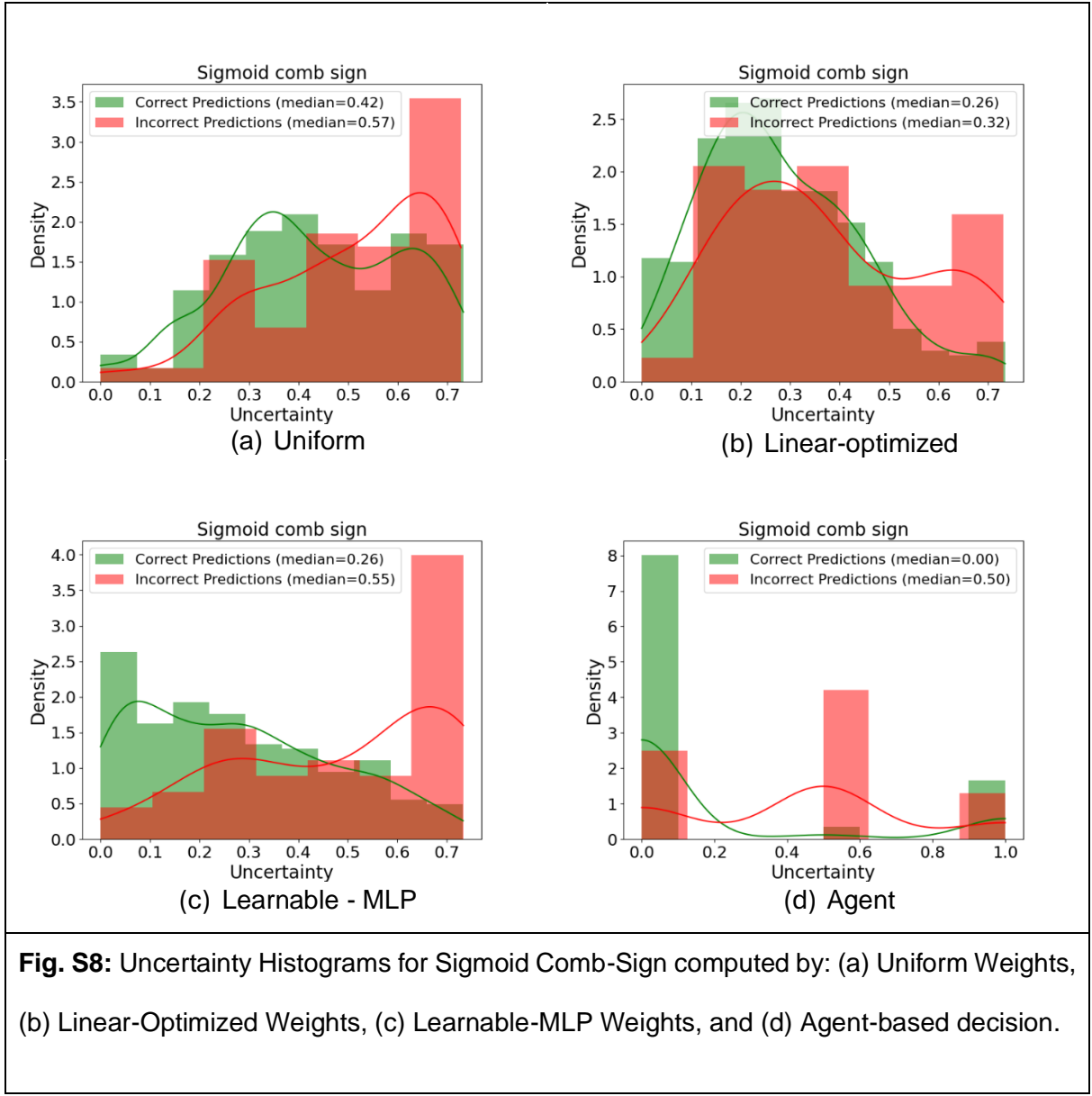

**Fig. S8:** Uncertainty Histograms for Sigmoid Comb-Sign computed by: (a) Uniform Weights, (b) Linear-Optimized Weights, (c) Learnable-MLP Weights, and (d) Agent-based decision.